\documentclass[final]{cvpr}
\usepackage{enumitem}
\usepackage{times}
\usepackage{epsfig}
\usepackage{graphicx}
\usepackage{amsmath}
\usepackage{amssymb}
\usepackage{makecell}
\usepackage{booktabs}
\usepackage[table,xcdraw]{xcolor}
\usepackage{multirow}
\usepackage{bbm}

\usepackage[pagebackref=true,breaklinks=true,colorlinks,bookmarks=false]{hyperref}

\def\cvprPaperID{4107} % *** Enter the CVPR Paper ID here
\def\confYear{CVPR 2021}

\begin{document}

\title{Railroad is not a Train: Saliency as Pseudo-pixel Supervision \\ for Weakly Supervised Semantic Segmentation}

\author{Seungho Lee\thanks{ indicates an equal contribution.}\\
Yonsei University\\
{\tt\small seungholee@yonsei.ac.kr}
\and
Minhyun Lee\footnotemark[1]\\
Yonsei University\\
{\tt\small lmh315@yonsei.ac.kr}
\and 
Jongwuk Lee\\
Sungkyunkwan University\\
{\tt\small jongwuklee@skku.edu}
\and 
Hyunjung Shim\thanks{Hyunjung Shim is a corresponding author.}\\
Yonsei University\\
{\tt\small kateshim@yonsei.ac.kr}
}

\maketitle
\thispagestyle{empty}
\pagestyle{empty}

%%%%%%%%% ABSTRACT

\begin{abstract}
Existing studies in weakly-supervised semantic segmentation (WSSS) using image-level weak supervision have several limitations: sparse object coverage, inaccurate object boundaries, and co-occurring pixels from non-target objects. To overcome these challenges, we propose a novel framework, namely Explicit Pseudo-pixel Supervision (EPS), which learns from pixel-level feedback by combining two weak supervisions; the image-level label provides the object identity via the localization map and the saliency map from the off-the-shelf saliency detection model offers rich boundaries. We devise a joint training strategy to fully utilize the complementary relationship between both information. Our method can obtain accurate object boundaries and discard co-occurring pixels, thereby significantly improving the quality of pseudo-masks. Experimental results show that the proposed method remarkably outperforms existing methods by resolving key challenges of WSSS and achieves the new state-of-the-art performance on both PASCAL VOC 2012 and MS COCO 2014 datasets. The code is available at \href{https://github.com/halbielee/EPS}{https://github.com/halbielee/EPS}.
\end{abstract}

\section{Introduction}

Weakly-supervised semantic segmentation (WSSS) utilizes weak supervision (\eg, image-level labels~\cite{pathak2015constrained, pinheiro2015image}, scribbles~\cite{lin2016scribblesup}, or bounding boxes~\cite{khoreva2017simple}) and aims at achieving competitive performances to the fully-supervised model, which requires pixel-level labels. Most existing studies adopt image-level labels as the weak supervision of the segmentation model. The overall pipeline of WSSS consists of two stages. Firstly, pseudo-masks are generated for target objects using an image classifier. Then, the segmentation model is trained using the pseudo-masks as supervision. The prevalent technique for generating pseudo-masks is class activation mapping (CAM)~\cite{zhou2016learning}, which provides object localization maps corresponding to their image-level labels. Because of the supervision gap between the fully (\ie, pixel-level annotations) and weakly (\ie, image-level labels) supervised semantic segmentation, WSSS has the following key challenges: 1) the localization map only captures a small fraction of target objects~\cite{zhou2016learning}, 2) it suffers from the boundary mismatch of the objects~\cite{kim2017two}, and 3) it hardly separates co-occurring pixels from target objects (\eg, the railroad from a train)~\cite{kolesnikov2016seed}. 

\begin{figure}[t]
\centering
\includegraphics[width=8 cm]{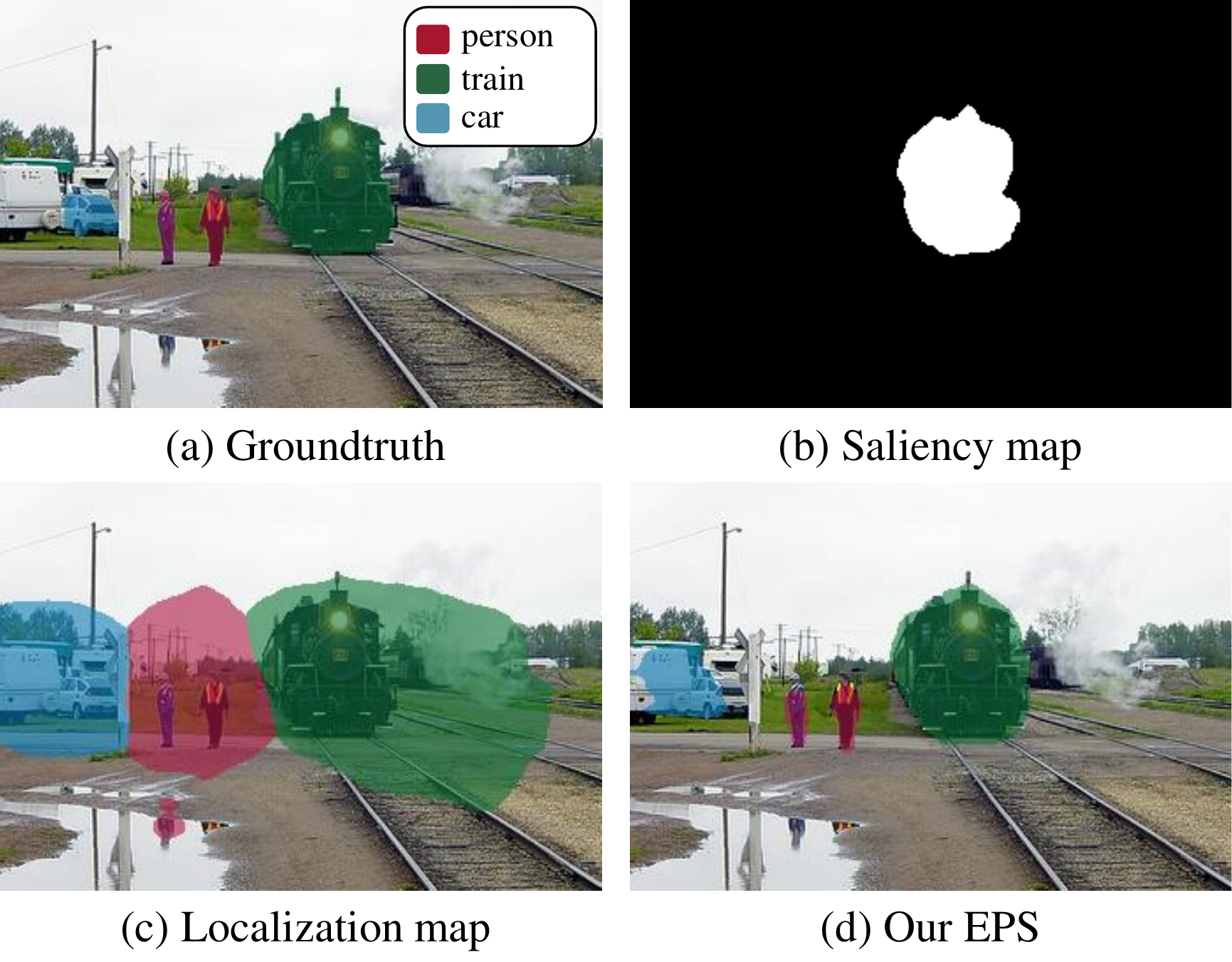}
\caption{Motivating example of utilizing both the saliency map and the localization map for WSSS. (a) Groundtruth, (b) saliency map via PFAN~\cite{zhao2019pyramid}, (c) localization map via CAM~\cite{zhou2016learning} and (d) our EPS utilizing both the saliency map and the localization map for training a classifier. Note that the saliency map cannot capture \emph{person} and \emph{car} while our result can correctly restore them, and the localization map overly captures two objects.} \vspace{-2mm}
\label{fig:concept}
\end{figure}

To address these problems, existing studies can be categorized into three pillars. The first approach expands object coverage to capture the full extent of objects by erasing pixels~\cite{choe2020attention,kim2017two, li2018tell}, ensembling score maps~\cite{jiang2019integral, lee2019ficklenet}, or using self-supervised signal~\cite{wang2020self}. However, they fail to determine accurate object boundaries of the target object because they have no clue to guide the object's shape. The second approach focuses on improving the object boundaries of pseudo-masks~\cite{fan2020learning,chen2020boundary}. Since they effectively learn object boundaries, they naturally expand pseudo-masks until boundaries. However, they still fail to distinguish coincident pixels of non-target objects from a target object. It is because the strong correlation between the foreground and the background (\ie, co-occurrence) is almost indistinguishable from an inductive bias (\ie, the frequency of observing the target object and its coincident pixels), as shown in~\cite{choe2020evaluating}. Lastly, the third approach aims to mitigate the co-occurrence problem using extra groundtruth masks~\cite{BMVC2016_92}, or the saliency map~\cite{oh2017exploiting, yao2020saliency}. However, \cite{BMVC2016_92,li2018tell} require strong pixel-level annotations, which are far from a weakly supervised learning paradigm. \cite{oh2017exploiting} is sensitive to the errors of the saliency map. Also, \cite{yao2020saliency} does not cover the full extent of objects and suffers from the boundary mismatch.

In this paper, our goal is to overcome the three challenges of WSSS by fully utilizing both the localization map (\ie, CAM from the image classifier trained with image-level labels) and the saliency map (\ie, the output of the off-the-shelf saliency detection model~\cite{hou2017deeply,nguyen2019deepusps,zhao2019pyramid}). We focus on a complementary relationship in the localization map and the saliency map. As illustrated in Figure~\ref{fig:concept}, the localization map can distinguish different objects but does not separate their boundaries effectively. Contrarily, while the saliency map provides rich boundary information, it does not reveal object identity. In this sense, we argue that our method using two complementary pieces of information can resolve the performance bottleneck of WSSS.

To this end, we propose a novel framework for WSSS, called \emph{Explicit Pseudo-pixel Supervision (EPS)}. To fully utilize the saliency map (\ie, both the foreground and the background), we design a classifier to predict $C+1$ classes, consisting of $C$ target classes and the background class. We leverage $C$ localization maps and the background localization map to estimate a saliency map. Then, the saliency loss is defined as the pixel-wise difference between the saliency map and our estimated saliency map. By introducing the saliency loss, the model can be supervised by pseudo-pixel feedback across all classes. We also use the multi-label classification loss to predict image-level labels. Therefore, we train the classifier to optimize both the saliency loss and the multi-label classification loss, synergizing the predictions for both the background and foreground pixels-- we find that our strategy can improve both the saliency map (Section~\ref{section3.3} and Figure~\ref{fig:sal}) and the pseudo-mask (Section~\ref{section:5.1} and Figure~\ref{fig:ablation}).

We stress that, because the saliency loss penalizes boundary mismatches via pseudo-pixel feedback, it can enforce our method to learn the object's accurate boundaries. As a byproduct, we can also capture the entire object by expanding the map until the boundaries. Because the saliency loss helps separate the foreground (\eg, a train) from the background, our method can assign the co-occurring pixels (\eg, a railroad) to the background class. Experimental results show that our EPS achieves remarkable segmentation performances, recording new state-of-the-art accuracies on PASCAL VOC 2012 and MS COCO 2014 datasets.

\section{Related Work}

\noindent\textbf{Weakly-supervised semantic segmentation.}
The general pipeline of WSSS is to generate pseudo-masks from a classification network and to use the pseudo-masks as supervision to train a segmentation network. Due to the scarcity of boundary information in the image-level label, many existing methods suffer from inaccurate pseudo-masks. To address this problem, cross-image affinity~\cite{fan2020cian}, knowledge graph~\cite{liu2020leveraging} and contrastive optimization~\cite{sun2020mining, zhang2020splitting} are used to improve the quality of pseudo-masks. \cite{chang2020weakly} proposes a self-supervised task to discover sub-categories to enforce the classifier to improve CAM. \cite{ahn2019weakly, ahn2018learning} implicitly exploit the boundary information by calculating affinities between pixels. \cite{zhang2020reliability} focuses on producing reliable pixel-level annotations and designs an end-to-end network for generating segmentation maps. \cite{huang2018weakly, kolesnikov2016seed} train the segmentation network by utilizing a boundary loss. Recently, \cite{araslanov2020single} uses a single segmentation-based model with a self-supervised training scheme. \cite{fan2020employing} focuses on the robustness of the segmentation network by utilizing multiple incomplete pseudo-masks.

\begin{figure*}[t]
\centering
\includegraphics[width=16cm]{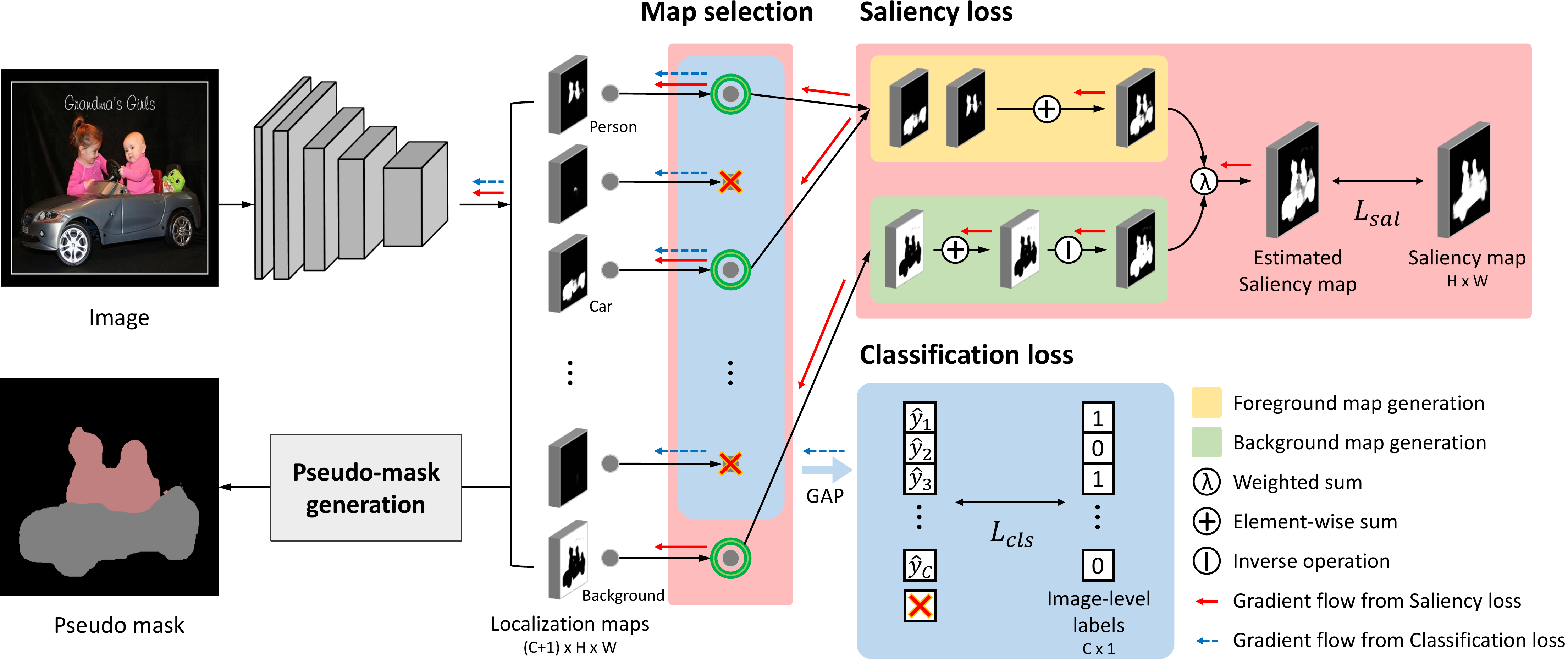}
\caption{The overall framework of our EPS. $C+1$ localization maps are generated from a backbone network. The actual saliency map is generated from the off-the-shelf saliency detection model. Some localization maps for target labels are selectively used to generate an estimated saliency map (Section~\ref{section3.2}). The overall framework is jointly trained with the saliency loss and the classification loss (Section~\ref{section3.3}). } \vspace{-2mm}
\label{fig:framework}
\end{figure*}

\vspace{1mm}
\noindent\textbf{Saliency-guided semantic segmentation.}
Saliency detection (SD) methods generate the saliency map that distinguishes between the foreground and the background in an image via external saliency datasets with pixel-level annotations~\cite{hou2017deeply, xiao2018deep, zhao2019pyramid}, or image-level annotations~\cite{wang2017learning}. Many WSSS methods~\cite{fan2020cian, huang2018weakly, lee2019ficklenet, li2018tell, wei2017object, wei2018revisiting} exploit the saliency map as the background cues of pseudo-masks. \cite{wei2016stc} utilizes the saliency map as the full supervision of single-object images. \cite{fan2018associating} uses an instance-level saliency map to learn the similarity graph for objects. \cite{chaudhry_dcsp_2017, wang2018weakly, yao2020saliency} combine saliency maps with class-specific attention cues to generate reliable pseudo-masks. \cite{zeng2019joint} jointly solves WSSS and SD using a single network to improve the performance of both tasks. Our EPS can be categorized into the saliency-guided method but is clearly distinguished from all others in the following reason. Most existing methods exploit the saliency map as a part of pseudo-masks or as the implicit guidance for refining the intermediate feature of the classifier. Contrarily, our method utilizes the saliency map as pseudo-pixel feedback for localization maps. Although \cite{zeng2019joint} is the most similar work to ours in the sense of utilizing two complementary information, they neither address the co-occurring problem nor handle the noisy saliency map issue.

\section{Proposed Method}

In this section, we propose a new framework for Weakly-supervised semantic segmentation (WSSS), called \emph{Explicit Pseudo-pixel Supervision (EPS)}. Considering two stages in WSSS, the first stage is to generate pseudo-masks and the second stage is to train the segmentation model. Here, our main contribution is to generate accurate pseudo-masks. Following the WSSS convention~\cite{fan2020learning,jiang2019integral,lee2019ficklenet,li2018tell,wang2020self,wei2017object}, we then train a segmentation model, where the generated pseudo-masks in the first stage are used as supervision.

\subsection{Motivation}
\label{section3.1}

Our key insight of EPS is to fully exploit two complementary information, \ie, the object identity from the localization map and boundary information from the saliency map. To this end, we utilize the saliency map as pseudo-pixel feedback to the localization map for both target labels and the background. We devise a classifier with an additional background class, leading to predict a total of $C+1$ classes, as shown in Figure~\ref{fig:framework}. Using the classifier, we can learn $C+1$ localization maps, \ie, $C$ localization maps for target labels and a background localization map.

We then explain how EPS can tackle both the boundary mismatch and co-occurrence problems in WSSS. To manage the boundary mismatch problem, we estimate the foreground map from $C$ localization maps and match it with the foreground of the saliency map. In this way, the localization maps for target labels can receive pseudo-pixel feedback from the saliency map, thereby improving the boundaries of objects. To mitigate the co-occurring pixels of non-target objects, we also match the localization map for the background with the saliency map. Since the localization map for the background also receives pseudo-pixel feedback from the saliency map, the co-occurring pixels can be successfully assigned to the background; the co-occurring pixels of non-target objects mostly overlap with the background. It is why our method can separate the co-occurring pixels from target objects. 

\begin{figure}[t]
\centering
\includegraphics[width=8cm]{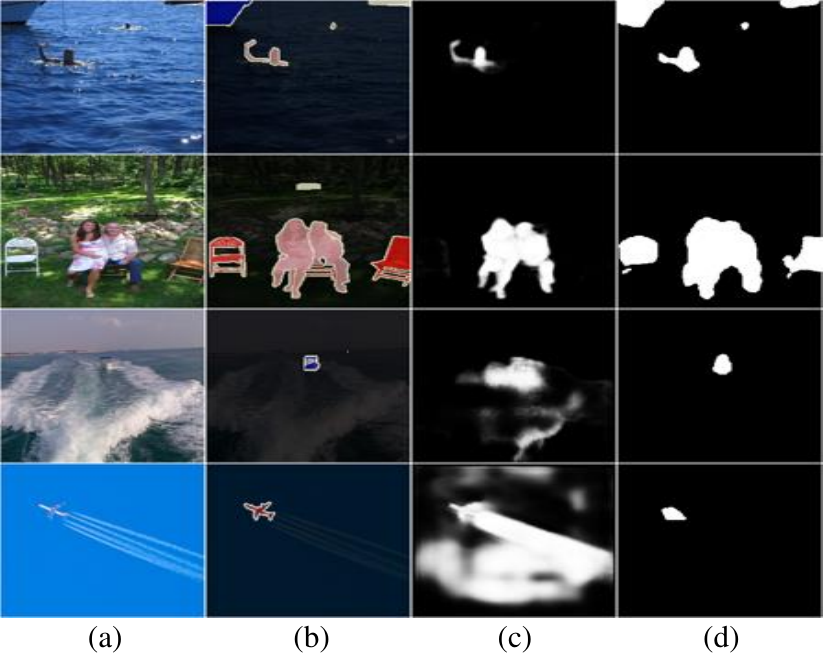}
\caption{Qualitative examples of estimated saliency maps on PASCAL VOC 2012. (a) Input images, (b) groundtruth, (c) saliency maps from \cite{zhao2019pyramid} and (d) our estimated saliency maps.} \vspace{-2mm}
\label{fig:sal}
\end{figure}

Lastly, the objective function of EPS is formulated with two parts: {the saliency loss} $\mathcal{L}_{sal}$ (marked by red box/arrow in Figure~\ref{fig:framework}) via the saliency map, and {the multi-label classification loss} $\mathcal{L}_{cls}$ (marked by blue box/arrow in Figure~\ref{fig:framework}) via image-level labels. By jointly training the two objectives, we can synergize the localization map and the saliency map with complementary information-- we observe that noisy and missing information of each other is complemented via our joint training strategy, as illustrated in Figure~\ref{fig:sal}. For example, the original saliency map obtained from the off-the-shelf model~\cite{hou2017deeply,nguyen2019deepusps,zhao2019pyramid} has missing and noisy information. On the other hand, our results successfully restore missing objects (\eg, boats or chairs) and remove the noise (\eg, water bubbles or contrail), which are evidently better than the original saliency map. Consequently, EPS can capture more accurate object boundaries and separate the co-occurring pixels from target objects. These advantages result in remarkable performance gains; Table~\ref{tab:seg_quan_voc_resnet101} reports that EPS remarkably outperforms existing models up to 3.8--10.6\% gains in terms of the segmentation accuracy.

\subsection{Explicit Pseudo-pixel Supervision}\label{section3.2}

We explain how to utilize the saliency map for pseudo-pixel supervision. The key advantage of the saliency map is to provide an object silhouette, which can better reveals object boundaries. To make use of this property, we match the saliency map with two cases: the foreground and the background. To make class-wise localization maps comparable with the saliency map, we merge the localization maps for target labels and generate a foreground map, $\mathbf{M}_{fg} \in \mathbb{R}^{H \times W}$. We can also represent the foreground by performing the inversion of a background map which is the localization map for the background label $\mathbf{M}_{bg} \in \mathbb{R}^{H \times W}$. (Later, we explain how to refine the foreground map to address noisy saliency maps.)

Specifically, we estimate the saliency map $\mathbf{\hat{M}}_{s}$ using $\mathbf{M}_{fg}$ and $\mathbf{M}_{bg}$ as follows:\vspace{-1mm}
\begin{equation}
\label{eq_esimate_sal}
{\small
\begin{split}
\mathbf{\hat{M}}_{s} = \lambda\mathbf{M}_{fg} + (1-\lambda)(1-\mathbf{M}_{bg}),
\end{split}}\vspace{-1mm}
\end{equation}
\noindent where $\lambda \in [0, 1]$ is a hyperparameter to adjust a weighted sum of the foreground map and the inversion of the background map. (By default, we set $\lambda$ to 0.5 in our experiments and an additional ablation study for $\lambda$ is found in the supplementary material.) Then, we define the saliency loss $\mathcal{L}_{sal}$ as the sum of pixel-wise differences between our estimated saliency map and an actual saliency map. (The formal definition of $\mathcal{L}_{sal}$ is presented in Section~\ref{section3.3}.)

It is worth noting that using the pre-trained model is regarded as weakly supervised learning, thus utilizing the saliency map has been widely accepted as a common practice in WSSS. Despite its popularity, adopting the fully supervised saliency detection model can be arguable in that they use pixel-level annotations from different datasets. In this paper, we investigate the effect of different saliency detection methods; 1) unsupervised and 2) fully supervised saliency detection models (see Section~\ref{section5.3}), and empirically show our method using any of them outperforms all other methods~\cite{fan2020learning,jiang2019integral,wang2018weakly, wei2016stc,yao2020saliency} using fully supervised saliency models. Whereas existing methods are limited to fully take advantage of the saliency map, our method incorporates the saliency map as pseudo-pixel supervision and exploits it as the cues for boundaries and co-occurring pixels.

\vspace{1mm}
\noindent\textbf{Map selection for handling saliency bias.} Previously, we assume that the foreground map can be the union of the localization maps for target labels; the background map can be the localization map of the background label. However, such a na\"ive selection rule may not be compatible with the saliency map computed by the off-the-shelf model. For example, the saliency map from \cite{zhao2019pyramid} often ignores some objects as salient objects (\eg, small people nearby a train in Figure~\ref{fig:concept}). This systematic error is inevitable because the saliency model learns the statistics of different datasets. Unless considering this error, the same error may propagate to our model and lead the performance degradation.

To tackle the systematic error, we develop an effective strategy using the overlapping ratio between the localization map and the saliency map. Specifically, the $i$-th localization map $\mathbf{M}_{i}$ is assigned to the foreground if $\mathbf{M}_{i}$ is overlapped with the saliency map more than $\tau$\%, otherwise the background. Formally, the foreground and the background map are computed by: \vspace{-1mm}
\begin{equation}
\label{eq_map_selection}
{\small
\begin{split}
&\mathbf{M}_{fg} = \sum_{i=1}^{C} {y_{i} \cdot \mathbf{M}_{i} \cdot \mathbbm{1}[\mathcal{O}(\mathbf{M}_i, \mathbf{M}_{s}) > \tau]}, \\
&\mathbf{M}_{bg} = \sum_{i=1}^{C} {y_{i} \cdot \mathbf{M}_{i} \cdot \mathbbm{1}[\mathcal{O}(\mathbf{M}_i, \mathbf{M}_{s}) \le \tau]} + \mathbf{M}_{C+1},
\end{split}}\vspace{-1mm}
\end{equation}
\noindent where $y \in \mathbb{R}^C$ is the binary image-level label and $\mathcal{O}(\mathbf{M}_i, \mathbf{M}_{s})$ is the function to compute the overlapping ratio between $\mathbf{M}_i$ and $\mathbf{M}_{s}$. For that, we first binarize the localization map and the saliency map such that: for pixel p, $\mathbf{B}_{k}(p) = 1$ if $\mathbf{M}_{k}(p) > 0.5$; $\mathbf{B}_{k}(p) = 0$, otherwise. $\mathbf{B}_{i}$ and $\mathbf{B}_{s}$ are the binarized maps corresponding to $\mathbf{M}_i$ and $\mathbf{M}_{s}$, respectively. We then compute the overlapping ratio between $\mathbf{M}_i$ and $\mathbf{M}_{s}$, \ie, $\mathcal{O}(\mathbf{M}_i ,\mathbf{M}_{s}) = |\mathbf{B}_i \cap \mathbf{B}_{s}| / |\mathbf{B}_{i}|$. We set $\tau=0.4$ regardless of datasets and backbone models. In the supplementary material, we show that our method is robust against the choice of $\tau$ (\ie, $\tau$ within [0.3, 0.5] shows the comparable performance).

Instead of a single localization map for the background label, we combine the localization map for the background label with the localization maps not selected as the foreground. Although it is simple, we can bypass the error of the saliency map and effectively train some objects neglected from the saliency map. (In Table~\ref{tab:strategy}, we report the effectiveness of the proposed strategy to overcome the error of the saliency map.)

\subsection{Joint Training Procedure}\label{section3.3}

Using the saliency map and image-level labels, the overall training objective of EPS consists of two parts, the saliency loss $\mathcal{L}_{sal}$ and the classification loss $\mathcal{L}_{cls}$. First, the saliency loss $\mathcal{L}_{sal}$ is formulated by measuring the average pixel-level distance between the actual saliency map $\mathbf{M}_{s}$ and the estimated saliency map $\mathbf{\hat{M}}_{s}$.
\vspace{-1mm}
\begin{equation}
{\small
\label{loss_sal}
\mathcal{L}_{sal} = \frac{1}{H\cdot W}||\mathbf{M}_{s}-\mathbf{\hat{M}}_{s}||^{2},
}
\vspace{-1mm}
\end{equation}
\noindent where $\mathbf{M}_{s}$ is obtained from the off-the-shelf saliency detection model-- PFAN~\cite{zhao2019pyramid} trained on DUTS dataset~\cite{wang2017learning}. Note that our method consistently outperforms all previous arts regardless of the saliency detection models.

Next, the classification loss is computed by a multi-label soft margin loss between the image-level label $y$ and its prediction $\hat{y} \in \mathbb{R}^C$, which is the result of the global average pooling on the localization map for each target class.
\vspace{-1mm}
\begin{equation}
{\small
\begin{split}
\label{loss_cls}
\mathcal{L}_{cls}= - \frac{1}{C} \sum_{i=1}^{C} y_i \log{\sigma(\hat{y_i})} + (1-y_i) \log{(1 - \sigma(\hat{y_i}))},
\end{split}}\vspace{-1mm}
\end{equation}
\noindent where $\sigma(\cdot)$ is the sigmoid function. Finally, the total training loss is the sum of the multi-label classification loss and the saliency loss, \ie, $\mathcal{L}_{total} = \mathcal{L}_{cls} + \mathcal{L}_{sal}$.

As shown in Figure~\ref{fig:framework}, $\mathcal{L}_{sal}$ is involved in updating the parameters of $C+1$ classes, including target objects and the background. Meanwhile, $\mathcal{L}_{cls}$ only evaluates the label prediction for $C$ classes, excluding the background class-- the gradient from $\mathcal{L}_{cls}$ does not flow into the background class. However, the prediction of the background class can be implicitly affected by $\mathcal{L}_{cls}$ because it supervises classifier training.

\section{Experimental Setup}
\noindent
\textbf{Datasets}. We conduct an empirical study on two popular benchmark datasets, PASCAL VOC 2012~\cite{everingham2015pascal} and MS COCO 2014~\cite{lin2014microsoft}. PASCAL VOC 2012 consists of 21 classes (\ie, 20 objects and the background) with 1,464, 1,449, and 1,456 images for training, validation, and test set, respectively. Following the common practice in semantic segmentation, we use the augmented training set with 10,582 images~\cite{hariharan2011semantic}. Next, COCO 2014 consists of 81 classes, including a background, with 82,081 and 40,137 images for training and validation, where images with no target classes are excluded as done in~\cite{choe2020attention}. Because the groundtruth segmentation labels of some objects overlap each other, we adopt the groundtruth segmentation labels from COCO-Stuff~\cite{caesar2018coco}, which solves the overlapping problem on the same COCO dataset.

\vspace{0.5mm}
\noindent
\textbf{Evaluation protocol}. We validate our method with the validation and the test set on PASCAL VOC 2012, and the validation set on COCO 2014. The evaluation results on the test set of PASCAL VOC 2012 is obtained from the official PASCAL VOC evaluation server. Also, we adopt mean intersection-over-union (mIoU) to measure the accuracy of segmentation models.

\vspace{0.5mm}
\noindent
\textbf{Implementation details}. We choose ResNet38~\cite{wu2019wider} as the backbone network of our method with the output stride of 8. All backbone models are pre-trained on ImageNet~\cite{deng2009imagenet}. We use the SGD optimizer with a batch size of 8. Our method is trained until 20k iterations with learning rate 0.01 (0.1 for the last convolutional layer). For data augmentation, we use a random scaling, random flipping, and random crop into $448 \times 448$. For the segmentation networks, we adopt DeepLab-LargeFOV (V1)~\cite{chen2014semantic} and DeepLab-ASPP (V2)~\cite{chen2017deeplab}, and VGG16 and ResNet101 for their backbone networks. Specifically, we use four segmentation networks: VGG16-based DeepLab-V1 and DeepLab-V2, ResNet101 based DeepLab-V1 and DeepLab-V2. More detailed setting is in the supplementary material.

\section{Experimental Results}

\subsection{Handling Boundary and Co-occurrence}\label{section:5.1}

\begin{figure}[t]
\centering
\includegraphics[width=8cm]{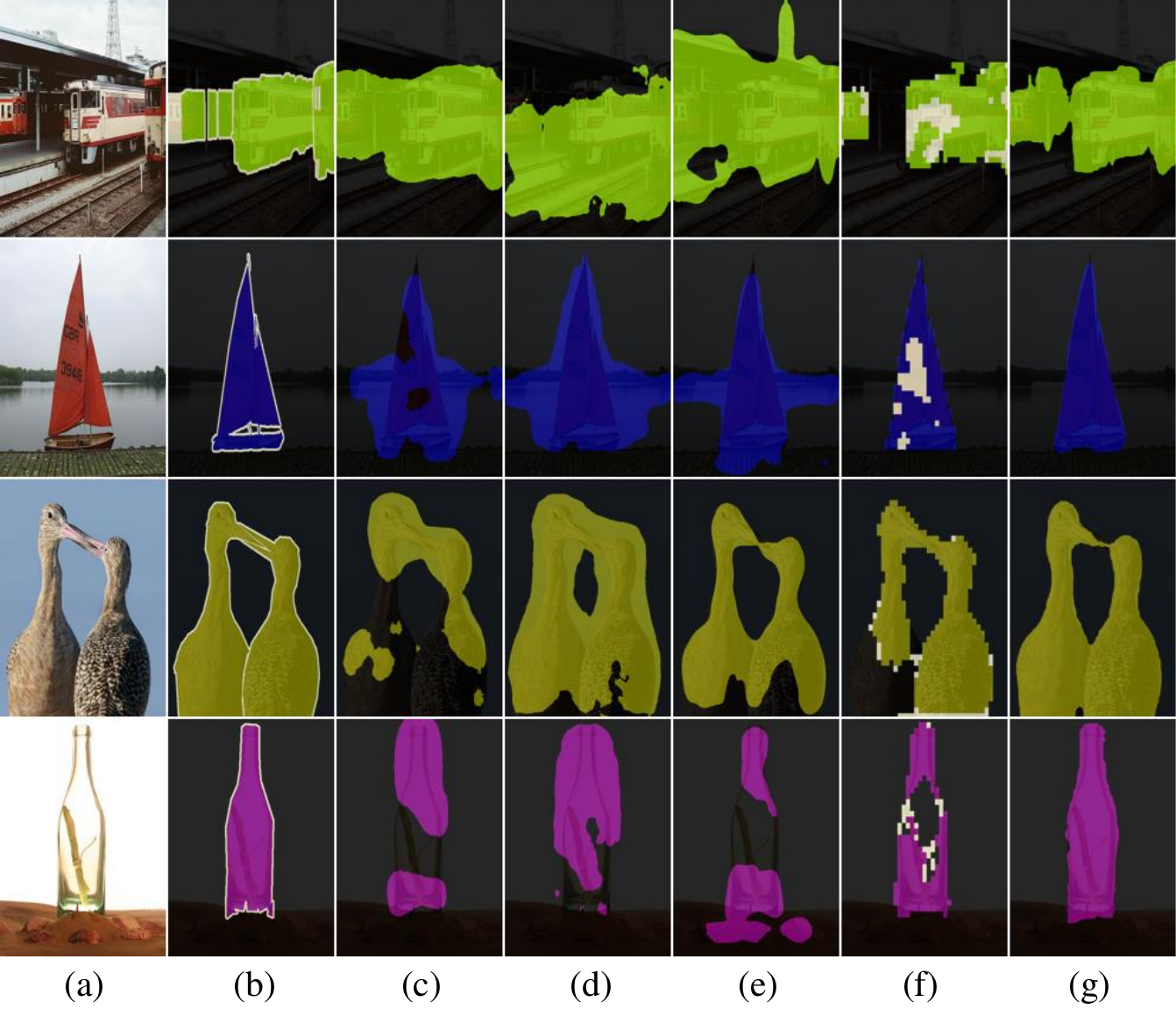}
\caption{Qualitative comparison for pseudo-masks on PASCAL VOC 2012. (a) Input images, (b) groundtruth, (c) CAM, (d) SEAM, (e) ICD, (f) SGAN and (g) our EPS.}
\label{fig:ablation} \vspace{-3mm}
\end{figure}

\noindent\textbf{Boundary mismatch problem.} To validate the boundary of pseudo-masks, we compare the quality of boundaries with the state-of-the-art methods~\cite{chen2020boundary, wang2020self, zhou2016learning}. We utilize SBD~\cite{hariharan2011semantic}, which provides boundary annotations and the boundary benchmark in PASCAL VOC 2011. As done in~\cite{chen2020boundary}, the quality of the boundary is evaluated in a class-agnostic manner by computing the edges of pseudo-masks from the Laplacian edge detector. Then, the boundary quality is evaluated by measuring recall, precision, and F1-score, comparing the predicted and groundtruth boundaries. Table~\ref{tab:boundary} reports that our method largely outperforms other methods in all three metrics. The qualitative examples in Figure~\ref{fig:ablation} show that our method can capture more accurate boundaries than all the other methods.

\begin{table}[]
\centering
{\small
\begin{tabular}{@{}lccc@{}}
\toprule
\multicolumn{1}{c}{Method}                      & Recall (\%) & Precision (\%) & F1-score (\%) \\ \midrule
\multicolumn{1}{l}{CAM~\cite{zhou2016learning}\textsubscript{CVPR'16}} & 22.3        & 35.8           & 27.5           \\
\multicolumn{1}{l}{SEAM~\cite{wang2020self}\textsubscript{CVPR'20}}    & 40.2        & 45.0           & 42.5           \\
\multicolumn{1}{l}{BES~\cite{chen2020boundary}\textsubscript{ECCV'20}} & 45.5        & 46.4           & 45.9           \\
\multicolumn{1}{l}{Our EPS}                        & 60.0        & 73.1          & 65.9           \\ \bottomrule
\end{tabular}
}
\vspace{2mm}
\caption{Boundary accuracy evaluated on the SBD trainval set. Note that the results of BES are measured from the boundary prediction network proposed in~\cite{chen2020boundary}.} \vspace{-2mm}
\label{tab:boundary}
\end{table}

\vspace{1mm}
\noindent \textbf{Co-occurrence problem.} As discussed in several studies~\cite{huang2018weakly, kolesnikov2016seed, li2018tell, oh2017exploiting}, we observe that some background classes frequently appear with target objects in PASCAL VOC 2012. We quantitatively analyze the frequency of co-occurred objects by employing the PASCAL-CONTEXT dataset~\cite{mottaghi2014role}, which provides pixel-level annotations for a whole scene (\eg,~\emph{water} and ~\emph{railroad}). We choose three co-occurring pairs; \emph{boat} with \emph{water}, \emph{train} with \emph{railroad}, and \emph{train} with \emph{platform}. We compare IoU for the target class and the \emph{confusion ratio} between a target class and its coincident class. The confusion ratio measures how much the coincident class is incorrectly predicted as the target class. The confusion ratio $m_{k,c}$ is calculated by $m_{k,c} = FP_{k,c}/TP_{c}$, where ${FP_{k,c}}$ is the number of pixels mis-classified as the target class $c$ for the coincident class $k$, and $TP_{c}$ is the number of true-positive pixels for the target class $c$. More detailed analysis on the co-occurrence problem is in the supplementary materials.

Table~\ref{tab:co_quantitative_v4} reports that EPS consistently shows a lower confusion ratio than other methods. SGAN~\cite{yao2020saliency} has quite a similar confusion ratio with ours, but our method captures the target class much accurately in terms of IoU. Interestingly, SEAM shows a high confusion ratio and even worse than CAM. It is because SEAM~\cite{wang2020self} learns to cover the full extent of target objects by applying self-supervised training, which is easily fooled by the coincident pixels of target objects. Meanwhile, CAM only captures the most discriminative region of target objects and does not cover the less discriminative parts, \eg, the coincident class. We can also observe this phenomenon in Figure~\ref{fig:ablation}.

% \usepackage{booktabs}

\newcommand{\R}{\textcolor{red}}
\newcommand{\B}{\textcolor{blue}}

\begin{table}[]
\centering
{\small
\begin{tabular}{@{}llll@{}}
\toprule
\multicolumn{1}{c}{\multirow{2}{*}{Method}} & \multicolumn{1}{c}{~\emph{boat} w/}   & \multicolumn{1}{c}{~\emph{train} w/}  & \multicolumn{1}{c}{~\emph{train} w/}  \\
& \multicolumn{1}{c}{~\emph{water}} & \multicolumn{1}{c}{~\emph{railroad}}          & \multicolumn{1}{c}{~\emph{platform}}  \\ \midrule
\multicolumn{1}{l}{CAM~\cite{zhou2016learning}\textsubscript{CVPR'16}}              & \B{0.74} (33.1)   & \B{0.11} (52.9)   & \multicolumn{1}{l}{\B{0.09} (49.6)}   \\
\multicolumn{1}{l}{SEAM~\cite{wang2020self}\textsubscript{CVPR'20}}                 & \B{1.13} (30.7)   & \B{0.24} (48.6)   & \multicolumn{1}{l}{\B{0.20} (45.5)}   \\
\multicolumn{1}{l}{ICD~\cite{fan2020learning}\textsubscript{CVPR'20}}               & \B{0.47} (41.4)   & \B{0.11} (56.7)   & \multicolumn{1}{l}{\B{0.09} (49.2)}   \\
\multicolumn{1}{l}{SGAN~\cite{yao2020saliency}\textsubscript{ACCESS'20}}            & \B{0.10} (42.3)   & \B{0.02} (48.8)   & \multicolumn{1}{l}{\B{0.01} (36.3)}   \\
\multicolumn{1}{l}{Our EPS}                                                         & \B{0.10} (55.0)   & \B{0.02} (78.1)   & \multicolumn{1}{l}{\B{0.01} (73.0)}   \\ \bottomrule
\end{tabular}
}
\vspace{2mm}
\caption{Comparison with representative existing methods handling the co-occurrence problem. Each entry is {$m_{k,c}$} in~\B{blue} (the lower the better) and IoU in the bracket (the higher the better).} \vspace{-2mm}
\label{tab:co_quantitative_v4}

\end{table}

\begin{table}[]
%\resizebox{\columnwidth}{!}{%
\centering

{\small
\begin{tabular}{@{}ccccc@{}}
\toprule
                            & Baseline  & Na\"ive   & Pre-defined   & Our adaptive \\ \midrule
\multicolumn{1}{l}{mIoU}    &66.1       & 66.5      & 67.9          & 69.4   \\ \bottomrule
\end{tabular}
}
\vspace{2mm}
\caption{Effect of map selection strategies. The accuracies of pseudo-masks using different map selection strategies are evaluated on the PASCAL VOC 2012 train set.} \vspace{-2mm}
\label{tab:strategy}
\end{table}

\subsection{Effect of Map Selection Strategies}
We evaluate the effectiveness of our map selection strategy to mitigate the error of the saliency map. We compare three different map selection strategies to the baseline, which does not use the map selection module. As the na\"ive strategy, the foreground map is the union of all object localization maps; the background map equals the localization map of the background class (\ie, na\"ive strategy). Next, we follow the na\"ive strategy with the following exceptions. The localization maps of several pre-determined classes (\eg, \emph{sofa}, \emph{chair}, and \emph{dining table}) are assigned to the background map (\ie, pre-defined class strategy). Lastly, the proposed selection method utilizes the overlapping ratio between the localization map and the saliency map, as explained in Section~\ref{section3.2} (\ie, our adaptive strategy).

    \begin{figure*}[t]
\centering
\includegraphics[width=17cm]{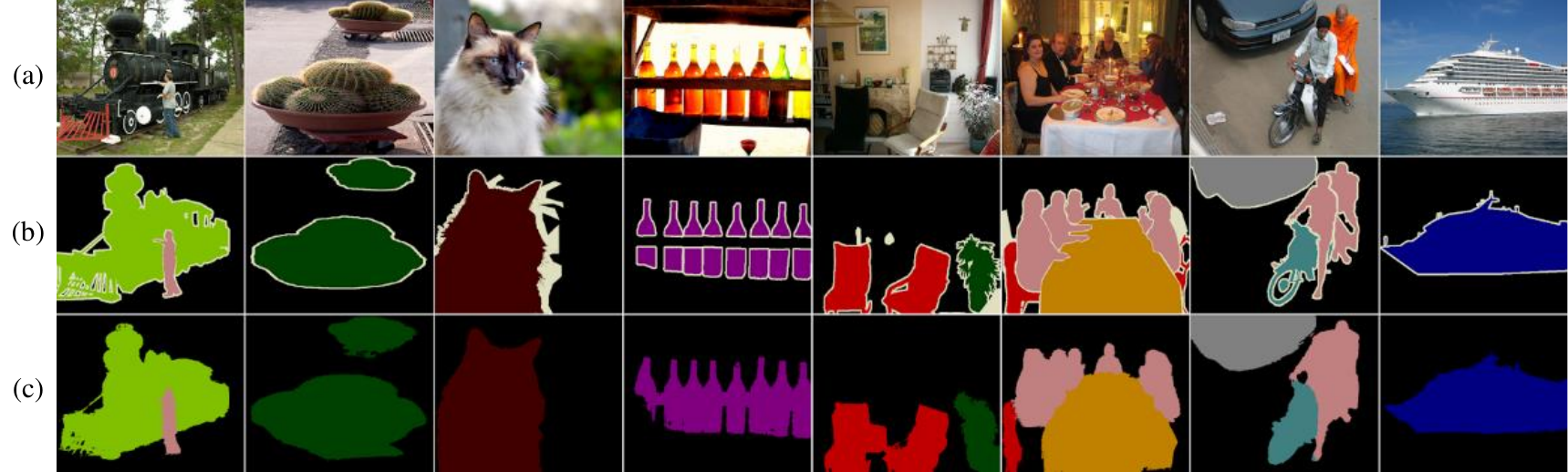}
\caption{Qualitative examples of segmentation results on PASCAL VOC 2012. (a) Input images, (b) groundtruth and (c) our EPS.}\vspace{-2mm}
\label{fig:seg_qual_voc}
\end{figure*}
\begin{table}[]
\centering
{\small
\begin{tabular}{@{}lccc@{}}
\toprule
\multicolumn{1}{c}{\multirow{2}{*}{Method}}         & w/o           & w/ &                                  w/ \\
                                                    & refinement    & CRF~\cite{krahenbuhl2011efficient}    & AffinityNet~\cite{ahn2018learning}    \\ \midrule
\multicolumn{1}{l}{CAM~\cite{zhou2016learning}\textsubscript{CVPR'16}}     & 48.0          & -                                     & 58.1                                  \\
\multicolumn{1}{l}{SEAM~\cite{wang2020self}\textsubscript{CVPR'20}}        & 55.4          & 56.8                                  & 63.6                                  \\
\multicolumn{1}{l}{ICD~\cite{chen2020boundary}\textsubscript{CVPR'20}*}     & 59.9          & 62.2                                  & -                                     \\
\multicolumn{1}{l}{SGAN~\cite{yao2020saliency}\textsubscript{ACCESS'20}*}     & 62.8          & -                                     & -                                     \\
\multicolumn{1}{l}{Our EPS}                            & 69.4          & 71.4                                  & 71.6                                  \\ \bottomrule
\end{tabular}
}
\vspace{2mm}
\caption{Accuracy (mIoU) of pseudo-masks evaluated on the PASCAL VOC 2012 train set. Note that * indicates that low-confident pixels are ignored; other methods use all pixels for evaluation.} \vspace{-3mm}
\label{tab:refinement}
\end{table}

Table~\ref{tab:strategy} shows that our adaptive strategy can effectively handle the systematic bias of the saliency map. The na\"ive strategy implies no bias consideration when generating the estimated saliency map from the localization maps. In this case, the performance of pseudo-masks is degraded, especially on \emph{sofa}, \emph{chair} or \emph{dining table} classes. The performance of using pre-defined classes shows that the bias can be mitigated by neglecting missing classes in the saliency map. However, as it requires manual selection by human observers, it is less practical and cannot make an optimal decision per image. Meanwhile, our adaptive strategy can handle the bias automatically and makes more effective decisions for a given saliency map. 

\subsection{Comparison with state-of-the-arts}
\label{section5.3}

\noindent \textbf{Accuracy of pseudo-masks.} We adopt a multi-scale inference by aggregating the prediction results from images with different scales, which is a common practice utilized in~\cite{ahn2018learning,wang2020self}. Then, We evaluate the accuracies of pseudo-masks in the train set by comparing our EPS with the baseline CAM~\cite{zhou2016learning} and three state-of-the-art methods, \ie, SEAM~\cite{wang2020self}, ICD~\cite{fan2020learning}, and SGAN~\cite{yao2020saliency}. Here, measuring the accuracy of the pseudo-masks in the train set is a common protocol in WSSS because the pseudo-masks of the train set are used to supervise the segmentation model. Table~\ref{tab:refinement} summarizes the accuracies of pseudo-masks and indicates that our method clearly outperforms all existing methods by large margins (\ie, 7--21\% gaps). Figure~\ref{fig:ablation} visualizes the qualitative examples of pseudo-masks, confirming that our method remarkably improves the object boundary and significantly outperforms three state-of-the-art methods in terms of the quality of pseudo-masks. Our method can capture the precise boundaries of objects (2nd row) and thus naturally cover the full extent of objects (3rd row), and also mitigate the coincident pixels (1st row). More examples and failure cases of our method are provided in the supplementary material.

\begin{table}[]
\normalsize
\centering
{\small
\begin{tabular}{@{}lccll@{}}
\toprule
\multicolumn{1}{c}{Method}                                                              & Seg.      & Sup.  & \multicolumn{1}{c}{val} & \multicolumn{1}{c}{test} \\ \midrule
\multicolumn{1}{l}{SEC~\cite{kolesnikov2016seed}\textsubscript{ECCV'16}}                & V1        & I.    & 50.7                    & 51.7                     \\
\multicolumn{1}{l}{AffinityNet~\cite{ahn2018learning}\textsubscript{CVPR'18}}           & V1        & I.    & 58.4                    & 60.5                     \\
\multicolumn{1}{l}{ICD~\cite{fan2020learning}\textsubscript{CVPR'20}}                   & V1        & I.    & 61.2                    & 60.9                     \\
\multicolumn{1}{l}{BES~\cite{chen2020boundary}\textsubscript{ECCV'20}}                  & V1        & I.    & 60.1                    & 61.1                     \\
\multicolumn{1}{l}{GAIN~\cite{li2018tell}\textsubscript{CVPR'18}}                       & V1        & I.+S. & 55.3                    & 56.8                     \\
\multicolumn{1}{l}{MCOF~\cite{wang2018weakly}\textsubscript{CVPR'18}}                   & V1        & I.+S. & 56.2                    & 57.6                     \\
\multicolumn{1}{l}{SSNet~\cite{zeng2019joint}\textsubscript{ICCV'19}}                   & V1        & I.+S. & 57.1                    & 58.6                     \\
\multicolumn{1}{l}{DSRG~\cite{huang2018weakly}\textsubscript{CVPR'18}}                  & V2        & I.+S. & 59.0                    & 60.4                     \\
\multicolumn{1}{l}{SeeNet~\cite{hou2018self}\textsubscript{NeurIPS'18}}                 & V1        & I.+S. & 61.1                    & 60.7                     \\
\multicolumn{1}{l}{MDC~\cite{wei2018revisiting}\textsubscript{CVPR'18}}                 & V1        & I.+S. & 60.4                    & 60.8                     \\
\multicolumn{1}{l}{FickleNet~\cite{lee2019ficklenet}\textsubscript{CVPR'18}}            & V2        & I.+S. & 61.2                    & 61.9                     \\
\multicolumn{1}{l}{OAA~\cite{jiang2019integral}\textsubscript{ICCV'19}}                 & V1        & I.+S. & 63.1                    & 62.8                     \\
\multicolumn{1}{l}{ICD~\cite{fan2020learning}\textsubscript{CVPR'20}}                   & V1        & I.+S. & 64.0                    & 63.9                     \\
\multicolumn{1}{l}{Multi-Est.~\cite{fan2020employing}\textsubscript{ECCV'20}}           & V1        & I.+S. & 64.6                    & 64.2                     \\
\multicolumn{1}{l}{Split. \& Merge.~\cite{zhang2020splitting}\textsubscript{ECCV'20}}   & V2        & I.+S. & 63.7                    & 64.5                     \\
\multicolumn{1}{l}{SGAN~\cite{yao2020saliency}\textsubscript{ACCESS'20}}                & V2        & I.+S. & 64.2                    & 65.0                     \\ \midrule
\multicolumn{1}{l}{\multirow{2}{*}{Our EPS}}                                            & V1        & I.+S. & 66.6                    & \textbf{67.9}            \\
\multicolumn{1}{l}{}                                                                    & V2        & I.+S. & \textbf{67.0}           & 67.3                     \\ \bottomrule

\end{tabular}
}
\vspace{2mm}
\caption{Segmentation results (mIoU) on PASCAL VOC 2012. All results are based on VGG16. The best score is in bold throughout all experiments.}\vspace{-3mm}
\label{tab:seg_quan_voc_vgg16}
\end{table}
\begin{table}[]
\normalsize
\centering
{\small
\begin{tabular}{@{}lccll@{}}
\toprule
\multicolumn{1}{c}{Method}                                                              & Seg.      & Sup.  & \multicolumn{1}{c}{val} & \multicolumn{1}{c}{test} \\ \midrule
\multicolumn{1}{l}{ICD~\cite{fan2020learning}\textsubscript{CVPR'20}}                   & V1        & I.    & 64.1                    & 64.3                     \\ 
\multicolumn{1}{l}{SC-CAM~\cite{chang2020weakly}\textsubscript{CVPR'20}}                & V1        & I.    & 66.1                    & 65.9                     \\
\multicolumn{1}{l}{BES~\cite{chen2020boundary}\textsubscript{ECCV'20}}                  & V2        & I.    & 65.7                    & 66.6                     \\
\multicolumn{1}{l}{LIID~\cite{liu2020leveraging}\textsubscript{TPAMI'20}}                  & V2        & I.    & 66.5                    & 67.5                     \\
\multicolumn{1}{l}{MCOF~\cite{wang2018weakly}\textsubscript{CVPR'18}}                   & V1        & I.+S. & 60.3                    & 61.2                     \\
\multicolumn{1}{l}{SeeNet~\cite{hou2018self}\textsubscript{NeurIPS'18}}                 & V1        & I.+S. & 63.1                    & 62.8                     \\
\multicolumn{1}{l}{DSRG~\cite{huang2018weakly}\textsubscript{CVPR'18}}                  & V2        & I.+S. & 61.4                    & 63.2                     \\
\multicolumn{1}{l}{FickleNet~\cite{lee2019ficklenet}\textsubscript{CVPR'18}}            & V2        & I.+S. & 64.9                    & 65.3                     \\
\multicolumn{1}{l}{OAA~\cite{jiang2019integral}\textsubscript{ICCV'19}}                 & V1        & I.+S. & 65.2                    & 66.4                     \\
\multicolumn{1}{l}{Multi-Est.~\cite{fan2020employing}\textsubscript{ECCV'19}}           & V1        & I.+S. & 67.2                    & 66.7                     \\
\multicolumn{1}{l}{MCIS~\cite{sun2020mining}\textsubscript{ECCV'20}}                    & V1        & I.+S. & 66.2                    & 66.9                     \\
\multicolumn{1}{l}{SGAN~\cite{yao2020saliency}\textsubscript{ACCESS'20}}                & V2        & I.+S. & 67.1                    & 67.2                     \\
\multicolumn{1}{l}{ICD~\cite{fan2020learning}\textsubscript{CVPR'20}}                   & V1        & I.+S. & 67.8                    & 68.0                     \\ \midrule
\multicolumn{1}{l}{\multirow{2}{*}{Our EPS}}                                            & V1        & I.+S. & \textbf{71.0}           & \textbf{71.8}            \\
\multicolumn{1}{l}{}                                                                    & V2        & I.+S. & 70.9                    & 70.8                     \\ \bottomrule
\end{tabular}
}
\vspace{2mm}
\caption{Segmentation results (mIoU) on PASCAL VOC 2012. All results are based on ResNet101.}\vspace{-2mm}
\label{tab:seg_quan_voc_resnet101}
\end{table}
\begin{figure*}[t]
\centering
\includegraphics[width=17cm]{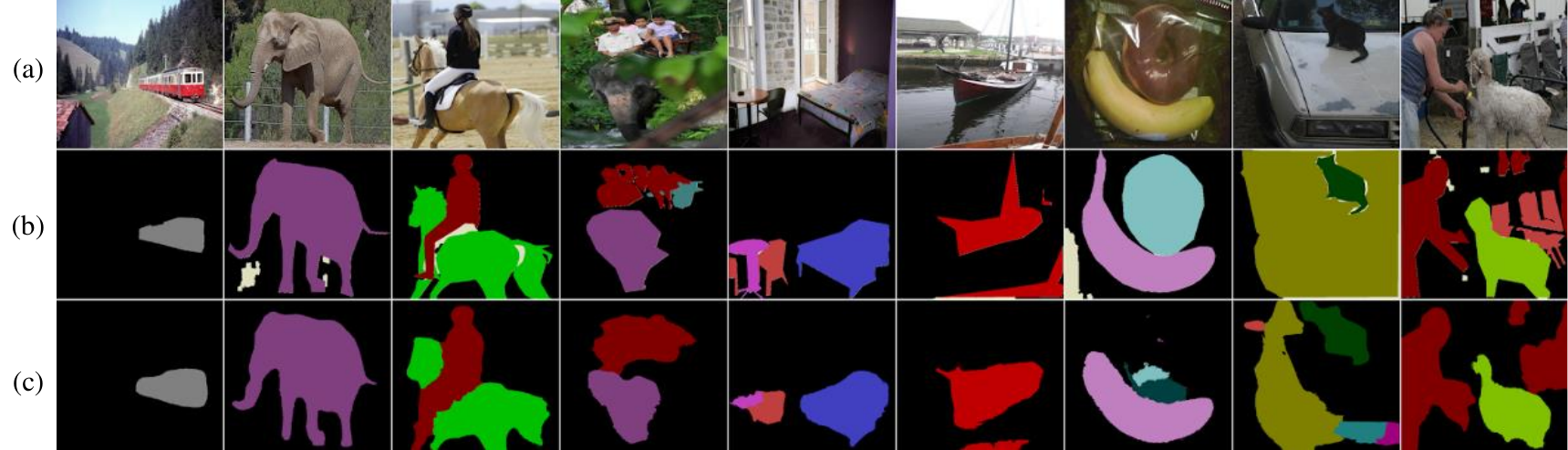}
\caption{Qualitative examples of segmentation results on MS COCO 2014. (a) Input images, (b) groundtruth and (c) our EPS.}\vspace{-2mm}
\label{fig:seg_qual_coco} 
\end{figure*}

\vspace{1mm}
\noindent \textbf{Accuracy of segmentation maps}. Previous methods~\cite{ahn2018learning, fan2020learning, wang2020self} generate pseudo-masks and refine them with the CRF post-processing algorithm~\cite{krahenbuhl2011efficient} or affinity network~\cite{ahn2018learning}. Meanwhile, as shown in Table~\ref{tab:refinement}, our generated pseudo-masks are accurate enough, thereby we train a segmentation network without any additional refinement for pseudo-masks. We extensively evaluate and precisely compare our method with others on the four segmentation networks in the Pascal VOC 2012 dataset.

% \usepackage{booktabs}
\begin{table}[]
\centering
{\small
\begin{tabular}{@{}lccc@{}}
\toprule
\multicolumn{1}{c}{Method}                                                  &Seg.       &Sup.   & \multicolumn{1}{c}{val}           \\ \midrule
\multicolumn{1}{l}{SEC~\cite{kolesnikov2016seed}\textsubscript{ECCV'16}}    & V1        &I.     & \multicolumn{1}{c}{22.4}          \\
\multicolumn{1}{l}{DSRG~\cite{huang2018weakly}\textsubscript{CVPR'18}}      & V2        &I.+S.  & \multicolumn{1}{c}{26.0}          \\
\multicolumn{1}{l}{ADL~\cite{choe2020attention}\textsubscript{TPAMI'20}}    & V1        &I.+S.  & \multicolumn{1}{c}{30.8}          \\
\multicolumn{1}{l}{SGAN~\cite{yao2020saliency}\textsubscript{ACESS'20}}     & V2        &I.+S.  & \multicolumn{1}{c}{33.6}          \\ \midrule
\multicolumn{1}{l}{Our EPS}                                                 & V2        &I.+S.  & \multicolumn{1}{c}{\textbf{35.7}} \\ \bottomrule
\end{tabular}
}
\vspace{2mm}
\caption{Segmentation results (mIoU) on MS COCO 2014. All results are based on VGG16.}\vspace{-2mm}
\label{tab:seg_quantitative_coco}
\end{table}

Our method performs remarkably better than other methods regardless of segmentation networks. Table~\ref{tab:seg_quan_voc_vgg16} reports that our method is more accurate than other methods with the same VGG16 backbone. Besides, our results on the VGG16 are comparable or even superior to other existing methods based on a more powerful backbone (\ie ResNet101 in Table~\ref{tab:seg_quan_voc_resnet101}). Our method also shows a clear improvement over existing methods. Finally, Table~\ref{tab:seg_quan_voc_resnet101} demonstrates that our method (under ResNet101 based DeepLab-V1 with saliency map) achieves the new state-of-the-art performance (71.0 for validation and 71.8 for test set) in the PASCAL VOC 2012 dataset. We highlight that the gains achieved by the existing state-of-the-art models were approximately 1\%. Meanwhile, our method achieves more than 3\% higher gains than the previous best record. Figure~\ref{fig:seg_qual_voc} visualizes the qualitative examples of our segmentation results on PASCAL VOC 2012. These results confirm that our method provides accurate boundaries and successfully resolves the co-occurrence problem.

In Table~\ref{tab:seg_quantitative_coco}, we further evaluate our method in the COCO 2014 dataset. We use VGG16 based DeepLab-V2 as the segmentation network to compare with SGAN~\cite{yao2020saliency}, which is the state-of-the-art WSSS model in the COCO dataset. Our method achieves 35.7 mIoU in the validation set, and it is 1.9\% higher than SGAN~\cite{yao2020saliency}. Consequently, we achieve the new state-of-the-art accuracy in the COCO 2014 dataset. These outstanding performances over the existing state-of-the-arts on both datasets confirm the effectiveness of our method; by fully utilizing both localization maps and the saliency map, it successfully captures the integral of target objects correctly and remedies the shortcomings of existing models. Figure~\ref{fig:seg_qual_coco} shows the qualitative examples of segmentation results on the COCO 2014 dataset. Our method performs well when a few objects appear without occlusions but less effective in handling many small objects. More examples and failure cases of our method are provided in the supplementary material.

\vspace{1mm}
\noindent \textbf{Effect of saliency detection models}. To investigate the effect of different saliency detection models, we adopt three saliency models; PFAN~\cite{zhao2019pyramid} (our default), DSS~\cite{hou2017deeply} used by OAA~\cite{jiang2019integral} and ICD~\cite{fan2020learning}, and USPS~\cite{nguyen2019deepusps} (\ie, the unsupervised detection model). The segmentation results (mIoU) under Resnet101 based DeepLab-V1 are 71.0/71.8 with PFAN, 70.0/70.1 with DSS, and 68.8/69.9 with USPS (validation set and test set), respectively. These scores support that our EPS using any of three different saliency models is still more accurate than all the other methods in Table~\ref{tab:seg_quan_voc_resnet101}. Notably, our EPS using the unsupervised saliency model outperforms all existing methods using the supervised saliency model.

\section{Conclusion}
We propose a novel weakly supervised segmentation framework, namely \emph{explicit pseudo-pixel supervision (EPS)}. Motivated by the complementary relationship between the localization map and the saliency map, our EPS learns from pseudo-pixel feedback combining with the saliency map and the localization map. Owing to our joint training scheme, we successfully complement noise or missing information on both sides. Consequently, our EPS can capture precise object boundaries and discard co-occurring pixels of non-target objects, remarkably improving the quality of pseudo-masks. Extensive evaluations and various case studies demonstrate the effectiveness of our EPS and the outstanding performances, the new state-of-the-art accuracies for WSSS on both PASCAL VOC 2012 and MS COCO 2014 datasets.

\noindent\textbf{Acknowledgements. }
We thank Duhyeon Bang and Junsuk Choe for the feedback. This research was supported by the Basic Science Research Program through the NRF Korea funded by the MSIP (NRF-2019R1A2C2006123, 2020R1A4A1016619), the IITP grant funded by the MSIT (2020-0-01361, Artificial Intelligence Graduate School Program (YONSEI UNIVERSITY)), and the Korea Medical Device Development Fund grant funded by the Korean government (Project Number:  202011D06).

{\small
\bibliographystyle{ieee_fullname}
\bibliography{egbib}
}

\end{document}